%% file: main.tex
\documentclass[conference, 9pt]{IEEEtran}

\IEEEoverridecommandlockouts
\usepackage{cite, amsmath, amsfonts, amssymb,amsthm,multicol,derivative,enumitem,graphicx,float,tikz,tikz-qtree,forest,verbatim,bm, lipsum, wasysym,afterpage,appendix, multirow, booktabs,placeins, dblfloatfix,subcaption}
\usepackage{algorithmic}
\usepackage{graphicx}
\usepackage{textcomp}
\usepackage{xcolor}

\usepackage{hyperref}
\hypersetup{draft}

\def\BibTeX{{\rm B\kern-.05em{\sc i\kern-.025em b}\kern-.08em
    T\kern-.1667em\lower.7ex\hbox{E}\kern-.125emX}}

\makeatletter
\newcommand{\linebreakand}{%
  \end{@IEEEauthorhalign}
  \hfill\mbox{}\par
  \mbox{}\hfill\begin{@IEEEauthorhalign}
}
\makeatother

\usepackage[utf8]{inputenc}
\usepackage[mathscr]{euscript}
\usepackage[ruled,linesnumbered]{algorithm2e}
\usetikzlibrary{fit,bayesnet, positioning, quotes,calc,shapes, arrows.meta}
\usepackage[numbers,sort&compress]{natbib}
\usepackage{hyperref}

\input{defs}

\begin{document}

\title{Deep Dynamic Probabilistic Canonical Correlation Analysis}

\author{\IEEEauthorblockN{Shiqin Tang}
\IEEEauthorblockA{\textit{Department of Data Science} \\
\textit{City University of Hong Kong}\\
Hong Kong, Country \\
t.sq@my.cityu.edu.hk}
\and
\IEEEauthorblockN{Shujian Yu}
\IEEEauthorblockA{\textit{Department of Artificial Intelligence} \\
\textit{Vrije Universiteit Amsterdam}\\
Amsterdam, Netherlands \\
s.yu3@vu.nl}
\and
\IEEEauthorblockN{Yining Dong}
\IEEEauthorblockA{\textit{Department of Data Science} \\
\textit{City University of Hong Kong}\\
Hong Kong \\
yining.dong@cityu.edu.hk}
\linebreakand 
\IEEEauthorblockN{S. Joe Qin}
\IEEEauthorblockA{\textit{Department of Computing and Decision Science} \\
\textit{Lingnan University}\\
Hong Kong \\
joeqin@lin.edu.hk}
}


\maketitle

\begin{abstract}
This paper presents Deep Dynamic Probabilistic Canonical Correlation Analysis (D$^2$PCCA), a model that integrates deep learning with probabilistic modeling to analyze nonlinear dynamical systems. Building on the probabilistic extensions of Canonical Correlation Analysis (CCA), D$^2$PCCA captures nonlinear latent dynamics and supports enhancements such as KL annealing for improved convergence and normalizing flows for a more flexible posterior approximation. D$^2$PCCA naturally extends to multiple observed variables, making it a versatile tool for encoding prior knowledge about sequential datasets and providing a probabilistic understanding of the system’s dynamics. Experimental validation on real financial datasets demonstrates the effectiveness of D$^2$PCCA and its extensions in capturing latent dynamics.
\end{abstract}

\section{INTRODUCTION}
\label{sec:intro}
In the quest to understand complex systems, Canonical Correlation Analysis (CCA) has emerged as an important tool for understanding the relationships between two sets of variables. 
This paper introduces Deep Dynamic Probabilistic Canonical Correlation Analysis (D$^2$PCCA), a novel contribution that integrates deep learning with probabilistic modeling to tackle the challenges of analyzing nonlinear dynamical system analysis.

CCA is a statistical method designed to explore the relationship between two sets of variables, denoted as $\bfx^1$ and $\bfx^2$. In its conventional setting, CCA seeks to obtain linear transformations of the variables so that the correlation of the transformed variables is maximized. 
Significant advancements have been made in the probabilistic modeling aspect of CCA. First introduced in~\cite{pcca_bach}, Probabilistic CCA (PCCA) assumes the existence of a shared latent factor $\mathbf{z}^0$ that generates both $\mathbf{x}^1$ and $\mathbf{x}^2$. PCCA was further generalized in~\cite{pcca_klami}, which posits the existence of additional latent factors, $\mathbf{z}^1$ and $\mathbf{z}^2$, uniquely associated with $\mathbf{x}^1$ and $\mathbf{x}^2$, respectively, alongside the common factor $\mathbf{z}^0$.
The Partial Least Squares (PLS) model is closely related to CCA in that both models aim to build connections between two datasets and explore their data generation mechanisms. 
However, they differ in that CCA adopts a more symmetric treatments on $\bfx^1$ and $\bfx^2$, while PLS assumes a directional or causal relationship between them; as a result, $\bfx^1$ and $\bfx^2$ can be perceived as input and output in the PLS setting. Notable implementation of Probabilistic PLS (PPLS) include generative approaches detailed in~\cite{ppls_bouhaddani, pcca_murphy} and a discriminative approach provided by~\cite{ppls_vidaurre}.

The past decade has seen the emergence of various dynamic variants of CCA designed to track the latent dynamics of two sets of variables, such as Dynamic-Inner PLS (DiPLS)~\cite{dicca} and Dynamic Probabilistic CCA (DPCCA)~\cite{dpcca}. 
This paper primarily explores the modeling framework of DPCCA, which extends the PCCA model~\cite{pcca_klami} into a dynamic setting.
The dynamic latent state in DPCCA is composed of three chains, with one shared by both observed variables, and each of the other two generating one of the observed variables. 

Our proposed Deep Dynamic Probabilistic CCA (D$^2$PCCA) model shares its graphical representation with DPCCA and uses a nonlinear structure similar to that of Deep Markov Models (DMMs)~\cite{dmm}, which are nonlinear generalization of linear dynamical systems (LDSs); DMMs preserves the first-order Markov property while replacing the linear Gaussian emission and transition functions with neural networks.
Like DPCCA, D$^2$PCCA is a generative approach and it makes no causal assumptions about the observed variables. 
However, training D$^2$PCCA as a deep latent variable model presents significant challenges, as traditional methods like the EM algorithm \cite{neal1998view} are not suitable due to the complexity and non-linearity of the model. To address this, we employ Amortized Variational Inference (AVI) \cite{avi, vi_blei}, which employs a shared inference network and uses Monte Carlo method for gradient estimation.

D$^2$PCCA is a versatile tool for capturing nonlinear latent dynamics in sequential data. 
It can accommodate various types of observations, whether continuous or discrete, by adjusting the emission network accordingly. 
By design, the dimensions of the D$^2$PCCA latent space can be smaller than those of the observations, compelling the model to focus on essential dynamics while filtering out systematic noise. 
D$^2$PCCA may be used as an intermediary step for further manipulation of the low-dimensional latent states extracted from the observations. 
Additionally, D$^2$PCCA provides a way to encode prior knowledge about temporal datasets, making it particularly effective when there is a clear separation in the data. For instance, in stock market analysis, D$^2$PCCA can track the evolution of stock prices across different sectors, uncovering both sector-specific dynamics and overarching market trends represented by a common latent factor.  

The paper is organized as follows: Section~\ref{sec:dpcca} introduces the mathematical formulation of Dynamic Probabilistic CCA (DPCCA), and Section~\ref{sec:d2pcca} proposes our neural network-enhanced D$^2$PCCA model. 
Section~\ref{sec:extions} details possible extensions that improve the performance or add features for D$^2$PCCA, including KL Annealing in training, normalizing flows for approximated posterior distribution, and multiple structural extensions. Experimental results on financial datasets are presented in Section \ref{sec:exp}.

\begin{figure}
\begin{center}
\begin{tabular}{cc}
\includegraphics[trim={.7cm 0cm 1.3cm 0cm}, clip, width=.217\textwidth]{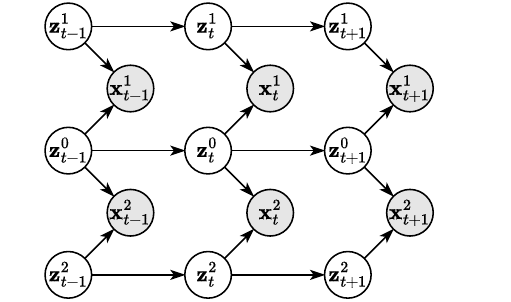} & \includegraphics[trim={.7cm 0cm 1.28cm 0cm}, clip, width=.22\textwidth]{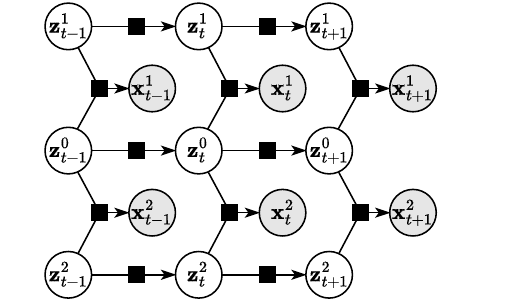}\\
(a) DPCCA & (b) D$^2$PCCA
\end{tabular}
\end{center}
\setlength{\abovecaptionskip}{0pt}  
\caption{Graphical models for DPCCA and D$^2$PCCA. The shaded nodes denote observed variables, while the unshaded ones denote latent variables. The arrows represent transition and emission models, and arrows with solid squares denote the usage of neural networks.}
\label{fig:dpcca}
\end{figure}

\section{Dynamic Probabilistic CCA (DPCCA)}
\label{sec:dpcca}
Dynamic Probabilistic CCA (DPCCA) is a natural extension to the static PCCA model proposed in~\cite{pcca_klami}. Let $\bfz_t^0$ be the latent variable underlying both observations, and let $\bfz_t^1$ and $\bfz_t^2$ be unique to $\bfx_t^1$ and $\bfx_t^2$ respectively. The transition and emission models of DPCCA are given by:
\begin{equation}
\begin{aligned}
p(\bfz_t^i|\bfz_{t-1}^i) &= \Normal(\bfz_t^i|A_i\bfz_{t-1}^i, V_i),\\
p(\bfx_t^j|\bfz_{t}^0,\bfz_{t}^j) &= \Normal(\bfx_t^j|W_j \bfz_t^0 + B_j \bfz_t^j, \sigma_j^2 I),
\end{aligned}
\label{eq:dpcca_model}    
\end{equation}
for $i \in \{0,1,2\}$ and $j \in \{1,2\}$. The graphical representation of the system can be seen in Fig.~\ref{fig:dpcca} (a). DPCCA can be viewed as Kalman filter with additional structural constraints. 
After reorganizing the variables and parameters as $\zhat_t = [\bfz_t^0, \bfz_t^1, \bfz_t^2]$, $\Ahat = \diag(A_0, A_1, A_2)$, $\xhat_t = [\bfx_t^1, \bfx_t^2]$, $W = [W_1^\top, W_2^\top]^\top$, and $B = \diag(B_1, B_2)$, the model specified in~\eqref{eq:dpcca_model} can be represented in a form compatible to that of a linear dynamical system:
\begin{equation}
\begin{aligned}
\zhat_t &= \Ahat \zhat_{t-1} + \bfu, &\bfu \sim \Normal(0, \hat{V}),\\
\xhat_t &= [W, B] \zhat_t + \bfep, &\bfep \sim \Normal(0, \hat{\Sigma}),
\end{aligned}
\label{eq:lds_model}   
\end{equation}
where
\begin{equation}
\begin{aligned}
\hat{V} = \begin{bmatrix}
    V_0 & & \\ &V_1& \\ && V_2
\end{bmatrix},\quad 
\hat{\Sigma} = \begin{bmatrix}
\sigma_1^2 I & \\ & \sigma_2^2 I
\end{bmatrix}.
\end{aligned}
\label{eq:lds_params}  
\end{equation}
Let $\theta = \{\Ahat, \hat{V}, W, B, \hat{\Sigma}\}$ contain the collection of model parameters. The joint distribution of the latent and observed states can be factorized as follows
\small
\begin{equation}
\begin{aligned}
p_\theta(\xhat_{1:T}, \zhat_{1:T}) &= \prod_{t=1}^T p_\theta(\zhat_t|\zhat_{t-1}) p_\theta(\xhat_t|\zhat_t) \\
&= \prod_{t=1}^T \Normal(\zhat_t|\Ahat \zhat_{t-1}, \hat{V}) \Normal(\xhat_t|[W, B] \zhat_t, \Sigmahat) .
\end{aligned}
\label{eq:lds_fact}  
\end{equation}
\normalsize
The parameters $\theta$ of the generative model can be learned using Expectation Maximization (EM) algorithm with objective function in the Maximization (M) step being
\footnotesize
\begin{equation}
\begin{aligned}
&\E_{p_\theta(\zhat_{1:T}|\xhat_{1:T})}[\log p_\theta(\xhat_{1:T}, \zhat_{1:T})]\\
\propto & -\frac{T}{2}\log |\Sigmahat| - \frac{1}{2}\sum_{t=1}^T \langle \tr((\xhat_t - [W, B] \zhat_t)(\xhat_t - [W, B] \zhat_t)^\top \Sigmahat^{-1} ) \rangle\\
&\quad -\frac{T}{2}\log |\Vhat| - \frac{1}{2}\sum_{t=1}^T \langle \tr((\zhat_t - \Ahat \zhat_{t-1})(\zhat_t - \Ahat \zhat_{t-1})^\top \Vhat^{-1} ) \rangle,
\end{aligned}
\label{eq:lds_obj}
\end{equation}
\normalsize
where the expectation is taken with respect to $p_\theta(\zhat_{1:T}|\xhat_{1:T})$, i.e.
\begin{equation}
\begin{aligned}
\langle \zhat_t \rangle &:= \E_{p_\theta(\zhat_t|\xhat_{1:T})}[\zhat_t],\\
\langle \zhat_t \zhat_{t}^\top  \rangle &:= \E_{p_\theta(\zhat_t|\xhat_{1:T})}[\zhat_t\zhat_{t}^\top],\\
\langle \zhat_t \zhat_{t-1}^\top  \rangle &:= \E_{p_\theta(\zhat_{t-1}, \zhat_t|\xhat_{1:T})}[\zhat_t\zhat_{t-1}^\top].
\end{aligned}
\label{eq:lds_exp}
\end{equation}
The calculation of the above expectations constitutes as the Expectation (E) step and can be obtained by Rauch-Tung-Striebel (RTS) smoother~\cite{sam}. EM algorithm involves iteratively updating $\theta$ towards minimizing the EM objective~\eqref{eq:lds_obj}, and updating the expectations in~\eqref{eq:lds_exp} until convergence.

\section{Deep Dynamic Probabilistic CCA (D$^2$PCCA)}
\label{sec:d2pcca}
Deep Dynamic Probabilistic CCA (D$^2$PCCA) is a nonlinear generalization of DPCCA with similar graphical structure as shown in Fig.~\ref{fig:dpcca}(b). Due to the nonlinearity of the emission and transition functions, the E step in the EM algorithm cannot be performed exactly and hence requires an inference network to approximate the posterior distribution of the latent states conditioned on the observations. In this section, we first introduce the generative network and then define the inference network, comparable to~\eqref{eq:dpcca_model} and~\eqref{eq:lds_exp} in the linear case, and details the training scheme on learning the parameters for both networks. For simplifying notations, we use $\MLP(h, f_1, f_2, \dots, f_n)$ to denote the output of a Multi-Layer Perceptron (MLP) of $n$ layers with activation function $f_i$ for each layer given an input $h$. Furthermore, we use $I$ to denote the identity function, and hence the layer with $I$ as activation function is a fully connected linear layer.  

The transition and emission models of D$^2$PCCA are given by:
\begin{equation}
\begin{aligned}
p(\bfz_t^i|\bfz_{t-1}^i) &= \Normal(\bfz_t^i|G_i(\bfz_{t-1}^i), \diag(S_i(\bfz_{t-1}^i))),\\
p(\bfx_t^j|\bfz_{t}^0,\bfz_{t}^j) &= \Normal(\bfx_t^j|M_j(\bfz_t^0, \bfz_t^j), \diag(V_j(\bfz_t^0, \bfz_t^j))),
\end{aligned}
\label{eq:d2pcca_model}    
\end{equation}
where $G_i$ and $S_i$ ($i=0,1,2$) form the gated transition network,
\begin{equation}
\begin{aligned}
h_t^i &= \MLP(\bfz_{t-1}^i, \ReLU, I),\\
g_t^i &= \MLP(\bfz_{t-1}^i, \ReLU, \sigma),\\
G_i(\bfz_{t-1}^i) &= g_t^i \odot h_t^i + (1-g_t^i) \odot \MLP(\bfz_{t-1}^i, I),\\
S_i(\bfz_{t-1}^i) &= \MLP(\ReLU(h_t^i), \mathrm{softplus}),
\end{aligned}
\label{eq:d2pcca_trans}    
\end{equation}
and $M_j$ and $V_j$ ($j=1,2$) form the emission network,
\begin{equation}
\begin{aligned}
h_{\bfx_t^j} &= \MLP([\bfz_t^0, \bfz_t^j], \ReLU, \ReLU),\\
M_j(\bfz_t^0, \bfz_t^j) &= \MLP(h_{\bfx_t^j}, I),\\
V_i(\bfz_t^0, \bfz_t^j) &= \exp(\MLP(h_{\bfx_t^j}, I)).
\end{aligned}
\label{eq:d2pcca_emit}    
\end{equation}
Proposed in~\cite{dmm}, the gated transition network shares a similarity with GRU as it trains a gate function $g_t^i$ to decide the portion of nonlinearity for $G_i$. We also implement an LSTM approach by training another gate function,
\begin{equation}
\begin{aligned}
w_t^i = \MLP(\bfz_{t-1}^i, \ReLU, \sigma),
\end{aligned}
\end{equation}
and redefining $G_i$ as
\begin{equation}
\begin{aligned}
G_i(\bfz_{t-1}) = g_t^i \odot h_t^i + w_t^i \odot \MLP(\bfz_{t-1}^i, I),
\end{aligned}
\end{equation}
and resulting training loss demonstrates no improvements. 
Defining $\xhat_t$ and $\zhat_t$ the same way as in the previous section, we can represent the generative network more compactly as
\begin{equation}
\begin{aligned}
p(\xhat_t|\zhat_t) &= p(\bfx_t^1|\bfz_t^0, \bfz_t^1) p(\bfx_t^2|\bfz_t^0, \bfz_t^2),\\
p(\zhat_t|\zhat_{t-1}) &\sim \Normal(\zhat_t| G(\zhat_{t-1}), \diag(S(\zhat_{t-1}))),
\end{aligned}
\end{equation}
where
\begin{equation}
\begin{aligned}
G(\zhat_{t-1}) = \begin{bmatrix}
G_0(\bfz_{t-1}) \\ G_1(\bfz_{t-1}^1) \\ G_2(\bfz_{t-1}^2) 
\end{bmatrix}, \quad S(\zhat_{t-1}) = \begin{bmatrix}
S_0(\bfz_{t-1}) \\ S_1(\bfz_{t-1}^1) \\ S_2(\bfz_{t-1}^2) 
\end{bmatrix}.
\end{aligned}
\end{equation}

Based on the graphical models in Fig.~\ref{fig:dpcca}, the three dynamic latent variables $\bfz_t^0$, $\bfz_t^1$, and $\bfz_t^2$ are independently distributed given previous states when the observed variables are unknown, and they become interdependent once the values of $\bfx^1$ and $\bfx^2$ are observed; such property enables us to treat the latent variables as a whole when performing inference. Since the latent states in D$^2$PCCA satisfies the first-order Markov property, the posterior distribution of the latent states can be factorized as
\begin{equation}
\begin{aligned}
p(\zhat_{1:T}|\xhat_{1:T}) &= p(\zhat_1|\xhat_{1:T})\prod_{t=2}^T p(\zhat_t|\zhat_{t-1}, \xhat_{t:T}).
\end{aligned}
\label{eq:post}
\end{equation}
Naturally, the proposal distribution $q$ chosen to approximate the true posterior distribution should have the same factorization form. In fact, the proposal distribution with the factorization form in~\eqref{eq:post} is referred to as structured model with information from the future (ST-R) in~\cite{dmm}. We train a backward RNN that takes $\xhat_{1:T}$ as input so that the RNN's $t$-th deterministic state, which we denote as $h_t^r$, contains information coming from the future, i.e. $\xhat_{t:T}$. As a result, the inference network is given by
\footnotesize
\begin{equation}
\begin{aligned}
q(\zhat_{1:T}|\xhat_{1:T}) &= \prod_{t=1}^T q(\zhat_t|\zhat_{t-1}, \xhat_{t:T})\\
&= \prod_{t=1}^T \Normal(\zhat_t|P(\zhat_{t-1}, h_t^r), \diag(Q(\zhat_{t-1}, h_t^r))),
\end{aligned}
\label{eq:st_l}
\end{equation}
\normalsize
where $P$ and $Q$ are referred to as combiner function in~\cite{dmm},
\begin{equation}
\begin{aligned}
&h_t^* = \frac{1}{2}\MLP(\zhat_{t-1},\tanh) + \frac{1}{2}h_t^r,\\
&P(\zhat_{t-1}, h_t^r) = \MLP(h_t^*, I),\\
&Q(\zhat_{t-1}, h_t^r) = \MLP(h_t^*, \mathrm{softplus}).
\end{aligned}
\label{eq:cbnr}
\end{equation}
Let $\theta$ and $\phi$ parameterize the generative and inference network respectively. The D$^2$PCCA model is trained by maximizing the variational lower bound (ELBO) given by
\begin{gather}
\begin{aligned}
\cL(\theta,\phi) &= \E_{q_\phi(\zhat_{1:T}|\xhat_{1:T})}\left[\log \frac{p_\theta(\xhat_{1:T}, \zhat_{1:T})}{q_\phi(\zhat_{1:T}|\xhat_{1:T})}\right]\\
&= \sum_{t=1}^T \E_{q_\phi(\zhat_t|\xhat_{t:T})}[\log p_\theta(\xhat_t|\zhat_t)]\\
&\qquad - \sum_{t=1}^T \mathop{\E\,[\KL(   q_\phi(\zhat_{t}|\zhat_{t-1},\xhat_{t:T})\|p_\theta(\zhat_t|\zhat_{t-1}))],}\limits_{q_\phi(\zhat_{t-1}|\xhat_{t-1:T})\qquad \qquad \qquad\qquad \qquad \, }
\end{aligned}
\label{eq:elbo_our}
\end{gather}
where $\KL$ denotes the Kullback-Leibler (KL) divergence defined as $\KL(q\|p) := \int_\cX q(\bfx) \log\frac{q(\bfx)}{p(\bfx)} \, \d\bfx$, where $p$ and $q$ are density functions with respect to the Lebesgue measure.


\section{D$^2$PCCA Extensions}
\label{sec:extions}
In this section, we details three extensions of D$^2$PCCA: KL-annealing in training, normalizing flows in inference, and related graphical models. 

\textbf{KL-Annealing}. Introduced in the context of Variational Autoencoders (VAEs) \cite{vae}, KL-annealing \cite{klanneal} is a training technique for preventing posterior collapse, a scenario in which the encoder fails to learn meaningful latent representations because the decoder is sufficiently powerful to model the data on its own. KL-annealing ameliorates posterior collapse by setting a smaller weight to the KL divergence term at the beginning of the training process and gradually raising the weight to one, which is effective because the KL divergence term is usually the dominant term initially. In the case of D$^2$PCCA, we can set $\beta$ to be a small positive value in the ELBO,
\begin{equation}
\begin{aligned}
\cL_\beta(\theta, \phi) &= \E_{q_\phi(\zhat_{1:T}|\xhat_{1:T})}[\log p_\theta(\xhat_{1:T}|\zhat_{1:T})]\\
&\qquad - \beta\KL(q_\phi(\zhat_{1:T}|\xhat_{1:T}) \| p_\theta(\zhat_{1:T})), 
\end{aligned}
\end{equation}
and gradually increase $\beta$ to one over a predefined number of epochs.

\begin{figure}
\centering
\begin{tabular}{@{}ccc@{}}
\includegraphics[trim={.7cm 0cm 1.3cm 0cm}, clip, width=.161\textwidth]{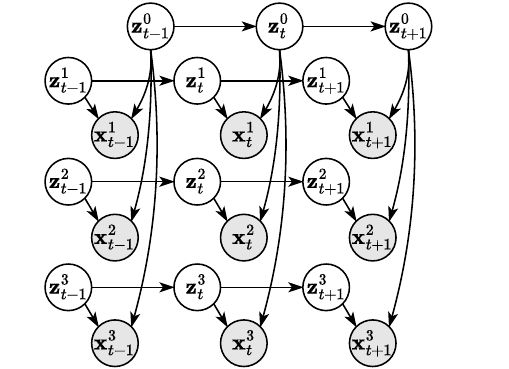}\hspace{-1.9em} & 
\includegraphics[trim={.7cm 0cm 2.1cm 0cm}, clip, width=.15\textwidth]{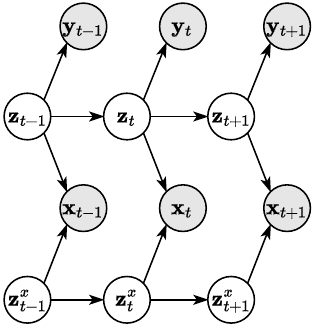}\hspace{-1.2em} & 
\includegraphics[trim={.7cm 0cm 1cm 0cm}, clip, width=.151\textwidth]{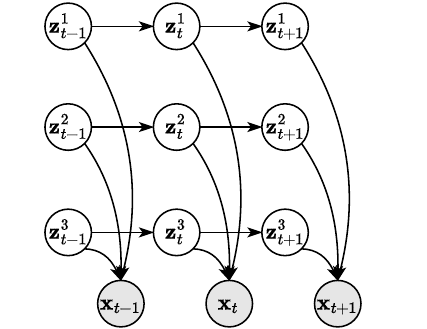}\\
(a) Multiset DPCCA & (b) DPPLS & (c) Factorial HMM
\end{tabular}
\caption{Graphical representations for (a) Multiset DPCCA, (b) DPPLS, and (c) Factorial HMM.}
\label{fig:dpcca_2}
\end{figure}

\textbf{Normalizing Flows}. The variational bound \ref{eq:elbo_our} is optimized for a given $\theta$ if $q_\phi(\zhat_{1:T}|\xhat_{1:T})$ approximates the true posterior distribution $p_\theta(\zhat_{1:T}|\xhat_{1:T})$ exactly. 
It is common practice to use an encoder network to parameterize tractable families of distributions, such as Gaussians; however, a variational family with a fixed structure often struggles to capture the complex characteristics of the true posterior distribution. 
Normalizing flows~\cite{nf, iaf} are neural network-based methods designed to approximate complex, multimodal density functions, and they have been applied to sequential modeling~\cite{NKF, sequence_AF}.

At each time step $t$, we first draw a sample from the base distribution, $\bfu_t \sim q_\phi(\bfu_t| \zhat_{t-1}, \xhat_{t:T})$, and then compute $\zhat_t = f_\omega(\bfu_t)$, where $f_\omega$ is a nonlinear bijection parameterized by $\omega$ with an inverse $g_\omega$. We have that the $\zhat_t$ satisfies distribution 
\begin{equation}
\begin{aligned}
q_\psi(\zhat_t|\zhat_{t-1}, \xhat_{t:T}) &= q_\phi(\bfu_t|\zhat_{t-1}, \xhat_{t:T}) \cdot |\det J(g_\omega) (\zhat_t)| \\
&= q_\phi(\bfu_t|\zhat_{t-1}, \xhat_{t:T}) \cdot |\det J(f_\omega) (\bfu_t)|^{-1},
\end{aligned}
\end{equation}
where $\psi = \{\phi, \omega\}$, and $J(g_\omega) (\zhat_t) = \mathrm{d}g_\omega(\zhat_t) \slash \mathrm{d} \zhat_t$ is the Jacobian.
As a result, the parameters of the generative and inference networks as well as the normalizing flows can be jointly trained by maximizing the updated variational lower bound given by
\small
\begin{equation}
\begin{aligned}
\cL(\theta,\phi,\omega) &= \E_{q_\psi(\zhat_{1:T}|\xhat_{1:T})}\Big[\log \frac{p_\theta(\xhat_{1:T}, \zhat_{1:T})}{q_\psi(\zhat_{1:T}|\xhat_{1:T})}\Big]\\
&= \sum_{t=1}^T \mathop{\E\,\big[\log p_\theta(\xhat_t|f_\omega(\bfu_t)) + \log p_\theta(f_\omega(\bfu_t)|f_\omega(\bfu_{t-1}))}\limits_{q_\phi(\bfu_{t-1}|\xhat_{t-1:T}) q_\phi(\bfu_{t}|f_\omega(\bfu_{t-1}), \xhat_{t:T})\qquad \qquad}\\
&\qquad - \log q_\phi(\bfu_t|f_\omega(\bfu_{t-1}), \xhat_{t:T}) + \log |\det J(f_\omega) (\bfu_t)|\big].
\end{aligned}
\label{eq:elbo_iaf}
\end{equation}
\normalsize


\textbf{Structural Extensions}. 
Proposed in~\cite{dpcca}, multiset DPCCA shown in Fig. \ref{fig:dpcca_2} (a) facilitates dynamic tracking of mutliple targets as each type of observed variable has a unique underlying factor and they also share a common latent factor.
The generative process of multiset DPCCA is a straightforward extension to that of DPCCA; in the former case, we let $i\in\{0, 1, \dots, D\}$ and $j \in \{1,\dots,D\}$ in \eqref{eq:dpcca_model}, where $D$ denotes the number of observations. 
D$^2$PCCA can easily extend to its multiset version. 
We propose the dynamic probabilistic PLS (DPPLS) in Fig. \ref{fig:dpcca_2} (b) as a dynamic version of the probabilistic PLS model mentioned in~\cite{pcca_murphy}.
DPPLS is applicable to scenarios where a subset of variables $\bfx_t$ has richer dynamics or significantly higher in dimensions compared to the other subset of variables $\bfy_t$, and the common latent factor is believed to sufficiently capture the the dynamics of $\bfy_t$.
D$^2$PPLS can be implemented with minor adjustments from dual-set D$^2$PCCA.
Factorial Hidden Markov Models (FHMMs) \cite{fhmm} in Fig. \ref{fig:dpcca_2} (c) model dynamic systems with multiple interrelated hidden processes. Each hidden chain reflects an aspect of the system's hidden state. Similar to DPCCA, the FHMMs' latent variables are interrelated conditioned on the past observations, but they have independent transition models. 
Due to limited space, this work focuses on multiset D$^2$PCCA in Section \ref{sec:exp}.

It may seem straightforward to extend the static probabilistic CCA (PCCA) model proposed by \cite{pcca_bach} into a deep dynamic version by modeling the joint probability distribution as
$$p(\bfx_t^1, \bfx_t^2, \bfz_t| \bfx_{t-1}^1, \bfx_{t-1}^2, \bfz_{t-1}) = p(\bfz_t|\bfz_{t-1}) p(\bfx_t^1|\bfz_t) p(\bfx_t^2|\bfz_t).$$
However, this extension becomes trivial if we confine the transition and emission models to the spherical Gaussian family, as the two observed variables can effectively be combined into a single entity, reducing the model to a standard deep Markov model.

\begin{figure}
\centering
\begin{tabular}{@{}cc@{}}
\includegraphics[trim={.8cm .8cm .2cm .8cm}, clip, width=.22\textwidth]{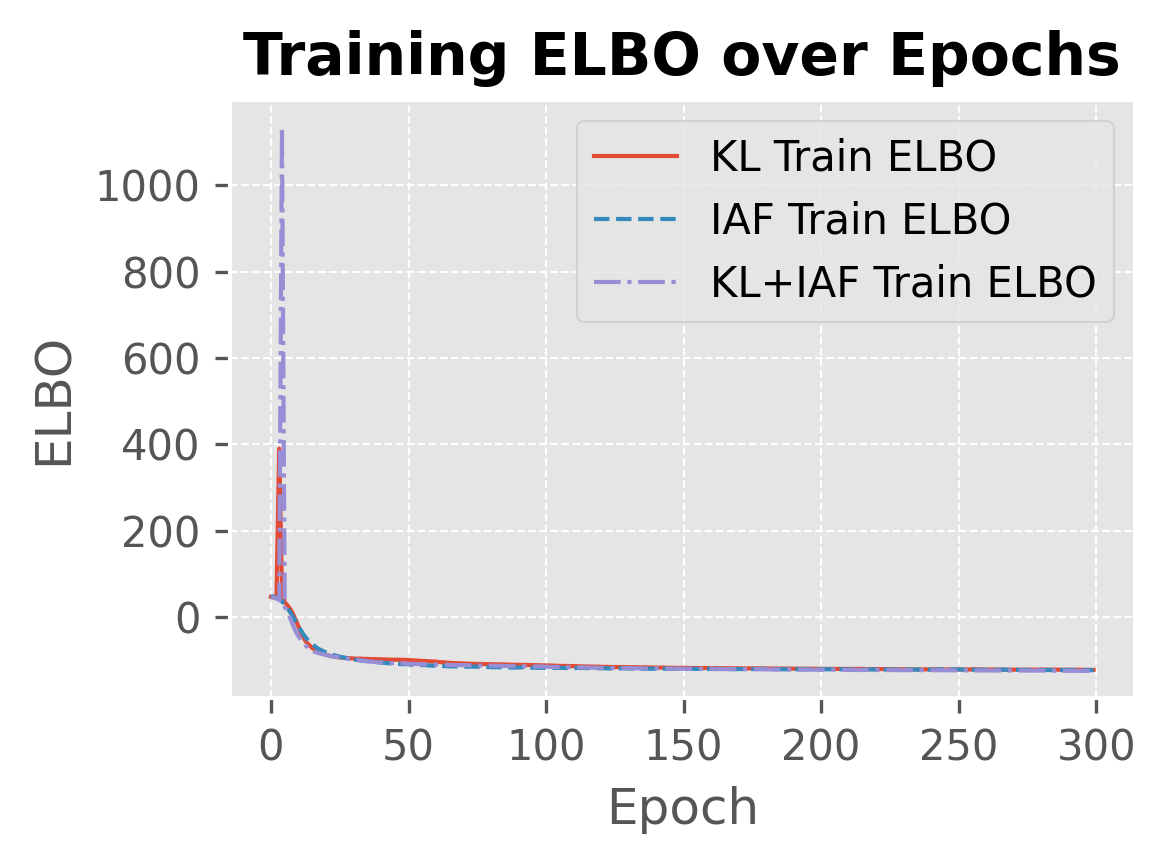}& \includegraphics[trim={1cm .8cm .2cm .8cm}, clip, width=.215\textwidth]{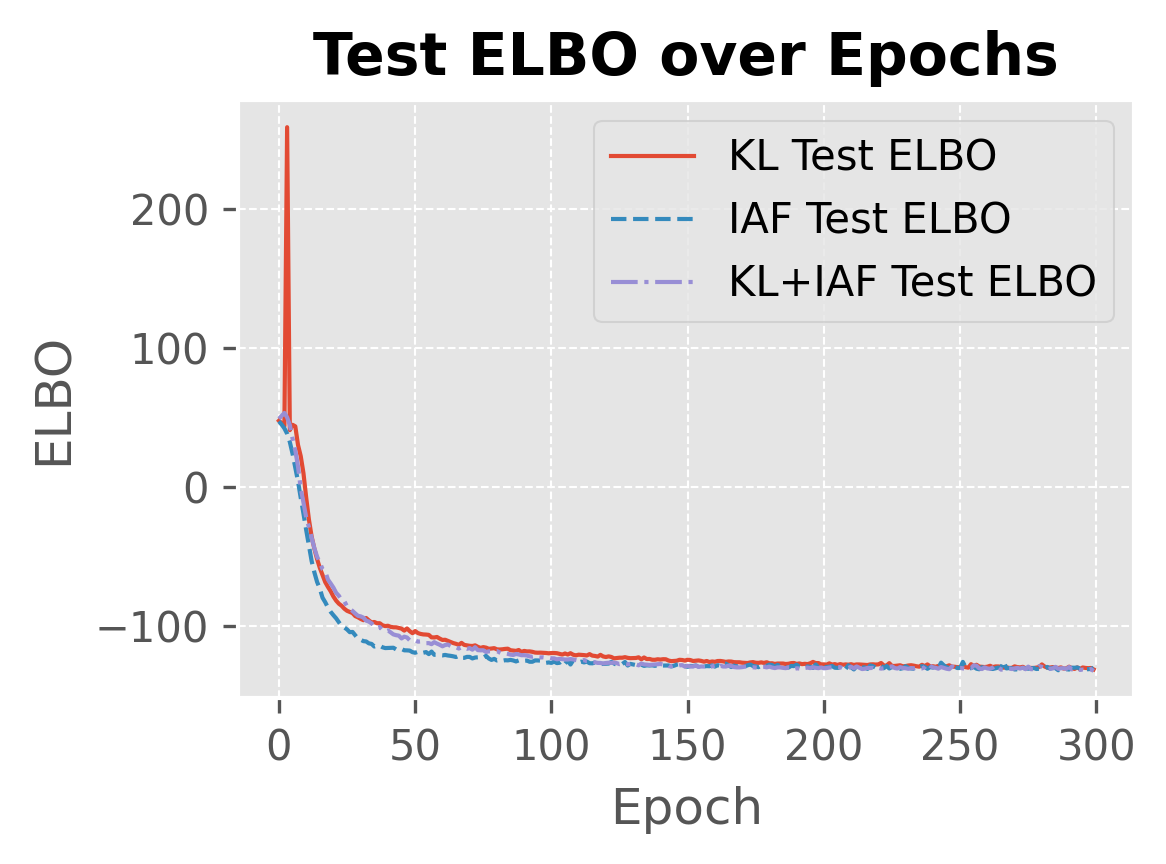}  
\end{tabular}
\setlength{\abovecaptionskip}{0pt}  
\caption{Convergence of ELBO during (left) training and (right) testing.}
\label{fig:elbo_tt}
\end{figure}

\section{Experiments}
\label{sec:exp}

In this section, we demonstrate the effectiveness of the multiset D$^2$PCCA in modeling nonlinear dynamical systems using a financial dataset. The model is trained using stochastic gradient descent with the ClippedAdam optimizer~\cite{adam}, utilizing the following parameters: a learning rate of $0.0003$, beta values of $0.96$ and $0.999$, a gradient clipping norm of $10$, and a weight decay rate of $2.0$. Our implementation is based on Pyro~\cite{pyro}, a probabilistic programming framework, and the code is available at {\url{https://github.com/marcusstang/D2PCCA}}.

We use daily closing prices of publicly traded companies on NASDAQ~\cite{kaggle} from 2018 to 2019, spanning a total of 503 trading days. From five financial sectors—finance, energy, technology, healthcare, and industrial—we select the 10 largest companies by market share within each sector. To prevent data overlap, we use the first $453$ data points for training and the remaining $50$ for testing. The training and test series are partitioned into sets of sequences using a sliding window of length $T = 30$ with a step size of $1$. Each sequence is centered to have zero mean and normalized using the standard deviation of the training series. We employ mini-batches of size $20$ during training. The training objectives on both the training and test sets over time are illustrated in Fig. \ref{fig:elbo_tt}. 

For the multiset D$^2$PCCA model, we set the dimension of the shared latent state to one and the dimension of each sector-specific factor to two. For models utilizing KL annealing, we initialize $\beta$ at $0.01$ and increase it linearly to $1$ over the first $100$ epochs. In the experiments with normalizing flows for posterior approximation, we employ five affine autoregressive flows, each with $70$ dimensions. The generative process of multiset DPCCA is implemented according to \eqref{eq:dpcca_model}, with the E-step in \eqref{eq:lds_exp} replaced by an inference network, similar to that used in D$^2$PCCA, to achieve faster convergence. 

We evaluate the performance of multiset DPCCA \cite{dpcca}, D$^2$PCCA, and its extensions in terms of ELBO, a lower bound on the model's log likelihood, as defined in \eqref{eq:elbo_our} and \eqref{eq:elbo_iaf}, and the Root Mean Square Error (RMSE) defined as:
\small
\begin{equation}
\begin{aligned}
\mathrm{RMSE} = \sqrt{\frac{1}{NT} \sum_{i=1}^N \sum_{t=1}^T \sum_{d=1}^D \|\bfx_{i,t}^d - \xhat_{i,t}^d\|^2 },
\end{aligned}
\label{eq:rmse}
\end{equation}
\normalsize
where $D=5$ represents the number of sectors, $N=20$ denotes the number of sequences in the test set, and $\xhat_{i,t}^d$ denotes the reconstructed observations.

As shown in TABLE \ref{tab:tab1}, D$^2$PCCA consistently outperforms its linear counterpart (DPCCA) in both ELBO and RMSE metrics. By utilizing nonlinear transition and emission models and allowing the variance to depend on the previous states rather than being fixed, D$^2$PCCA achieves a significantly higher ELBO, indicating a better fit to the data.
D$^2$PCCA, enhanced with KL annealing and a normalizing flow posterior, achieves the highest ELBO, which is also the training objective. However, it does not show a significant advantage over its competitors in RMSE. We believe this is because stock data contains simpler patterns compared to more complex signals like speech or video, and therefore, the dataset lacks sufficient nonlinear patterns for D$^2$PCCA to fully leverage its capabilities.

\begin{figure}
\begin{center}
\begin{tabular}{c}
\includegraphics[trim={0cm 0cm 0cm .8cm}, clip, width=.45\textwidth]{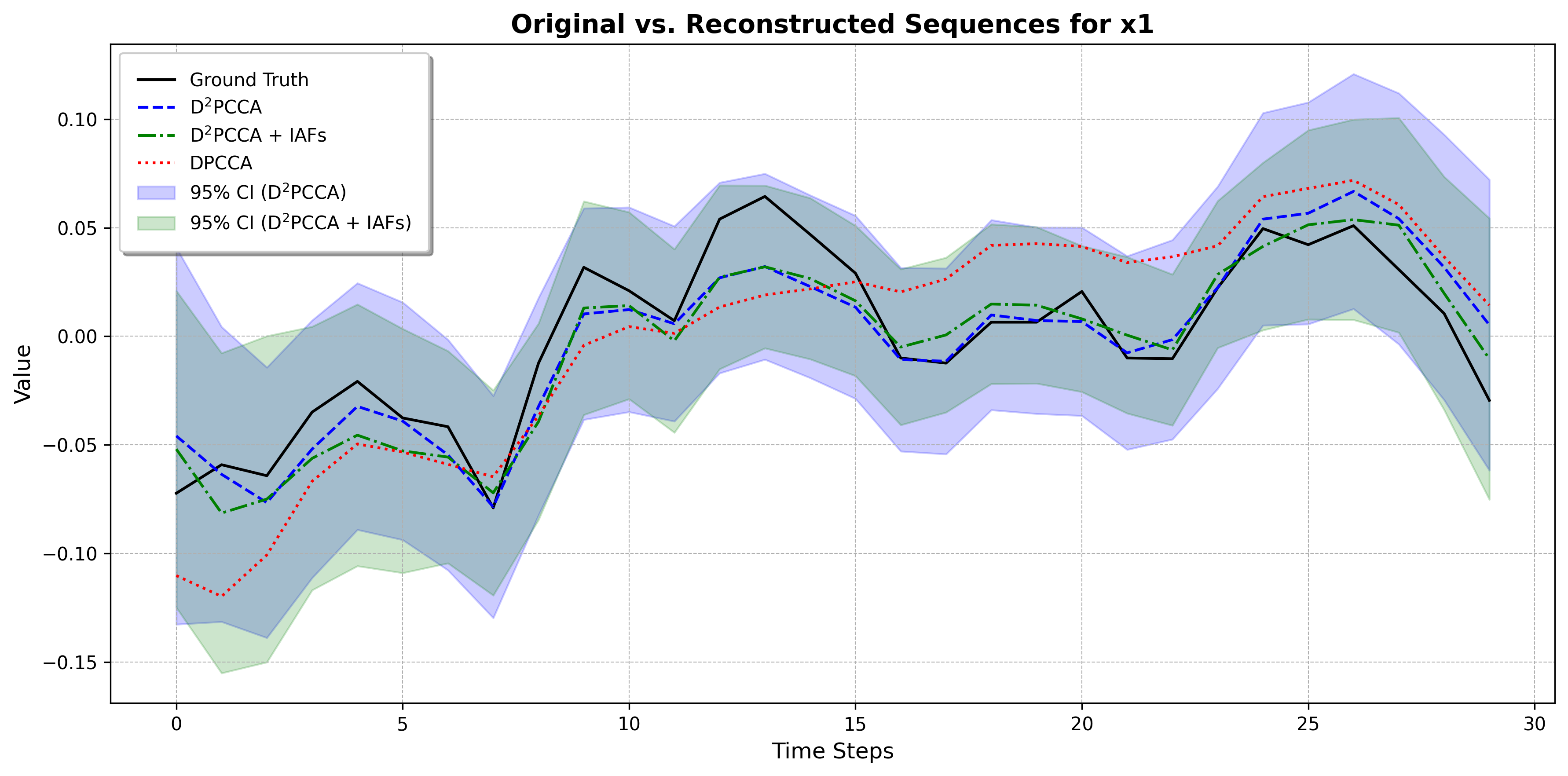} 
\end{tabular}
\end{center}
\setlength{\abovecaptionskip}{0pt}  
\caption{Comparison of model predictions against the ground truth over time. The shaded region denotes the confidence interval, $\mu \pm 1.96 \sigma$.}
\label{fig:recon}
\end{figure}

Fig. \ref{fig:recon} provides a qualitative assessment of the models' predictions. For the D$^2$PCCA models with an IAF posterior, we sample the latent states $\bfz_t$  because normalizing flows do not have an analytic mean, whereas for the other models, we use the expected value of the latent states, i.e. $\bfz_t = \E_{q(\bfz_t|\bfx_{1:T})}[\bfz_t]$. The error bounds for DPCCA are omitted because they are significantly larger than those of the D$^2$PCCA models, which partially explains DPCCA's lower likelihood (its variance is fixed rather than dynamically adjusted). 
We observe that the variance is much larger at the beginning of each sequence, likely due to the presence of a nonlinear trend in the stock data.


\begin{table}
\centering
\begin{tabular}{lcc}
\toprule
 & \textbf{Test ELBO $\uparrow$} & \textbf{Test RMSE $\downarrow$} \\
\midrule
\textbf{D$^2$PCCA} &  &  \\
\hspace{1em} + KL & 130.07 & \textbf{0.0179} \\
\hspace{1em} + IAF & 130.96 & 0.0183 \\
\hspace{1em} + KL + IAF & \textbf{131.27} & 0.0184 \\
\textbf{DPCCA} & 69.77 & 0.0181 \\
\bottomrule
\end{tabular}
\caption{
Performance evaluation of D$^2$PCCA and DPCCA, measured by the ELBO and reconstruction RMSE on the test set. ``KL" indicates training with KL annealing, while ``IAF" denotes the incorporation of normalizing flows in posterior approximation.
}
\label{tab:tab1}
\end{table}

\section{Conclusion \& Future Work}
\label{sec:conclusions}

In conclusion, this paper introduces D$^2$PCCA, a model that combines deep learning and probabilistic modeling to analyze the intricacies of nonlinear dynamical systems. The model's adaptability to various data types and its ability to encode prior knowledge about the datasets make it a versatile tool for a wide range of applications. 

In addition to conducting more experimental validations on different data types (e.g., audio and video), we can explore further variations of the model; for instance, we could augment the latent state transitions with a feedback connection from previous observations, as suggested in \cite{dmm}, or relax the first-order Markov assumptions, similar to the approaches in \cite{vrnn, dvae}.

\newpage
\label{sec:refs}
\bibliographystyle{IEEEtran} 
\bibliography{refs}

\end{document}

%% file: defs.tex
\theoremstyle{defn}

\theoremstyle{thm}

\theoremstyle{lem}

\theoremstyle{rmrk}

\theoremstyle{rslt}

\theoremstyle{expl}

\theoremstyle{cor}

\theoremstyle{smry}

\definecolor{lightblue}{RGB}{173,216,230} 
\definecolor{lightgray}{gray}{.97}

\newcommand{\MLP}{\mathrm{MLP}}
\newcommand{\ReLU}{\mathrm{ReLU}}

\let\oldforall\forall
\renewcommand{\forall}{\oldforall \, }
\let\oldLeftrightarrow\Leftrightarrow
\renewcommand{\Leftrightarrow}{\oldLeftrightarrow \,\, }
\let\oldexist\exists
\renewcommand{\exists}{\oldexist \: }

\newcommand{\KL}{D_{\mathrm{KL}}}

\newcommand{\diag}{\mathrm{diag}}
\newcommand{\tr}{\mathrm{tr}}

\newcommand{\Normal}{\mathcal{N}}

\newcommand{\cL}{\mathcal{L}} 
\newcommand{\E}{\mathbb{E}}
\newcommand{\xhat}{\hat{\mathbf{x}}}
\newcommand{\zhat}{\hat{\mathbf{z}}}

\newcommand{\Ahat}{\hat{A}}

\newcommand{\bfx}{\mathbf{x}}
\newcommand{\bfu}{\mathbf{u}}

\newcommand{\bfz}{\mathbf{z}}

\newcommand{\bfy}{\mathbf{y}}

\newcommand{\bfep}{\bm{\epsilon}}

\newcommand{\cX}{\mathcal{X}}

\newcommand{\Vhat}{\hat{V}}
\newcommand{\Sigmahat}{\hat{\Sigma}}

\renewcommand{\d}{\mathrm{d}}
